\documentclass{article}
\usepackage{ijcai20}

\usepackage{times}
\usepackage{soul}
\usepackage{url}
\usepackage[hidelinks]{hyperref}
\usepackage[utf8]{inputenc}
\usepackage[small]{caption}
\usepackage{graphicx}
\usepackage{amsmath}
\usepackage{amsthm}
\usepackage{booktabs}
\usepackage{algorithm}
\usepackage{algorithmic}
\usepackage{amssymb}

\title{Fine-grained Semantic Constraint in Image Synthesis}

\author{
	Pengyang Li$^1$\And
	Donghui Wang$^1$\footnote{Contact Author}\\
	\affiliations
	$^1$Zhejiang University\\
	\emails
	\{pyli, dhwang\}@zju.edu.cn
}

\begin{document}

\maketitle


\begin{abstract}
	
In this paper, we propose a multi-stage and high-resolution model for image synthesis that uses fine-grained attributes and masks as input. With a fine-grained attribute, the proposed model can detailedly constrain the features of the generated image through rich and fine-grained semantic information in the attribute. With mask as prior, the model in this paper is constrained so that the generated images conform to visual senses, which will reduce the unexpected diversity of samples generated from the generative adversarial network. This paper also proposes a scheme to improve the discriminator of the generative adversarial network by simultaneously discriminating the total image and sub-regions of the image. In addition, we propose a method for optimizing the labeled attribute in datasets, which reduces the manual labeling noise. 
Extensive quantitative results show that our image synthesis model generates more realistic images.

\end{abstract}


\begin{figure*}
	\begin{center}
		\includegraphics[width=0.9\linewidth]{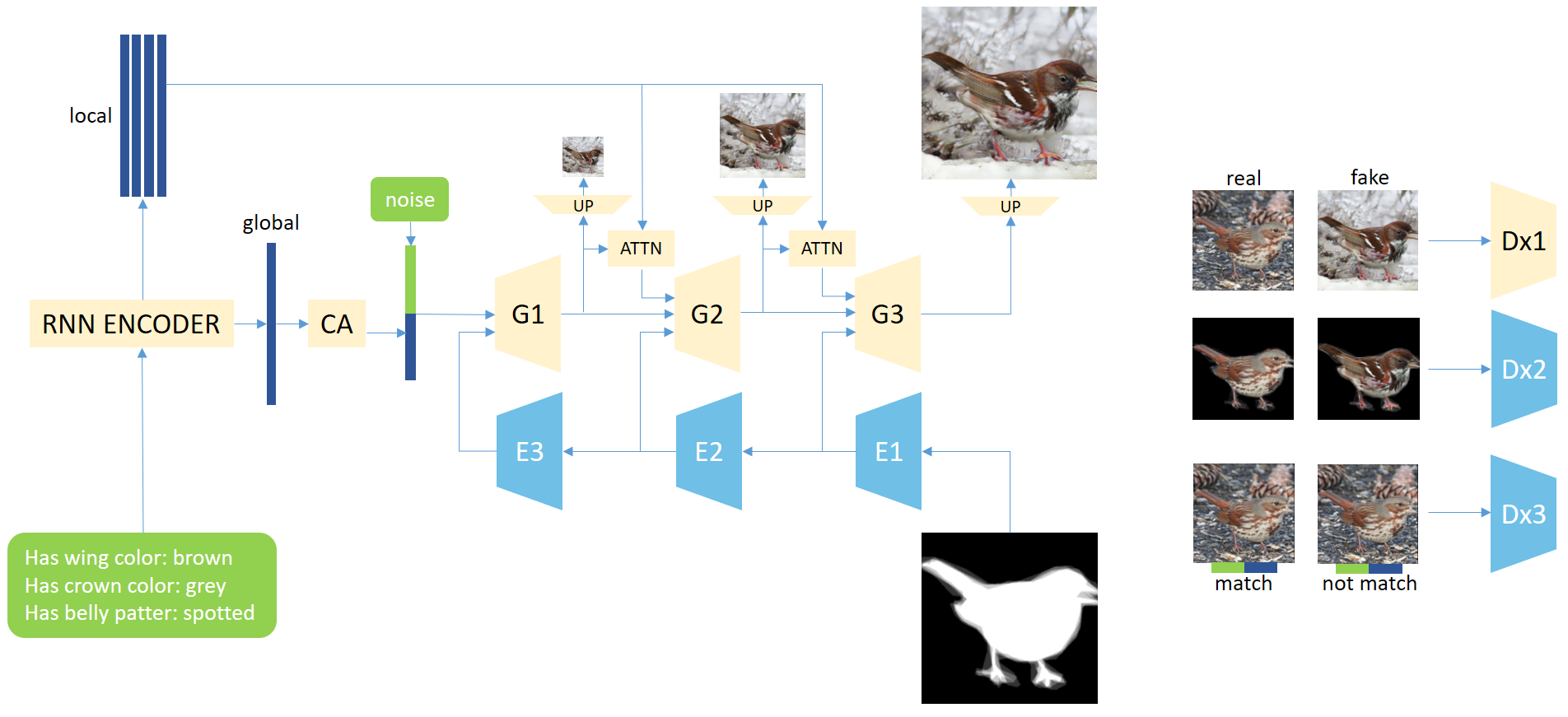}
	\end{center}
	\caption{Overall architecture.}
	\label{fig:arch}
\end{figure*}

\section{Introduction}

The research of image generation tasks shows the development of the current field of artificial intelligence. Ordinary people will have the opportunity to generate fictional images of different scenes through their own desires, through which people can feel the first-line artificial intelligence research and provide impetus for the further integration of artificial intelligence into society. Rencent works for the image generation task has shown that computers are fully capable of using deep networks to generate images that conform to certain semantic information, even enough to confuse real images. However, the capabilities of various image generation models for satisfying semantic constraints are not good enough at present, and the images generated by the current image generation model are still very limited in helping other visual tasks. AttnGAN\cite{xu2018attngan} is one of the great masters of image generation in recent years. It tries to generate details of bird images by fine-grained constraints. The result is that it can only constrain the rough part features, and the birds generated on the test set have low visual quality. The problem of image generation under fine-grained semantic constraints has become one of the focuses of image generation tasks. We hope to further constrain the details of images while generating more realistic images. 

This paper aims to explore further image generation algorithms under fine-grained semantic constraints. Based on the research of image generation methods with generating models in recent years, we have found a way to further solve the problem of image generation under fine-grained semantic constraints. Our specific research work and contributions are as follows: 

\noindent 1. Discover and solve the semantic granularity of image generation. At present, most of the image generation models take the embedded vector of a short sentence as a semantic input. However, the semantic information contained in a short sentence is limited and can be disturbed by irrelevant words. In order to obtain more rich, accurate and fine-grained semantic input, we re-select the embedded vector of the feature attribute as input. With this change, we can constrain the generated image with more fine-grained semantic information while maintaining the correspondence between semantics and generated images. 

\noindent 2. Optimize the accuracy of manually labeled attribute. Because we re-select the attribute as input, we need to accurately label the training data set with the attribute. This kind of data set is undoubtedly very rare, and the CUB\cite{wah2011caltech} dataset we chose to conveniently make comparison with existing image generation studies is undoubtedly not such a data set. Although the CUB dataset has been labeled with attributes, we have found through careful investigation that nearly 40\% of its attribute labels do not match the image. In this paper, a manually labeled attribute denoising algorithm is proposed to clean this inaccurately labeled data set, which improves the correspondence between the generated model semantics and the generated image. 

\noindent 3. Improve the visual quality of generated images. At present, most of the image generation models can generate high-quality images, but in general, there are always strange images that are incomprehensible in the results. To solve this problem, we added mask constraints in the image generation process. In addition, many image generation models tend to only make overall discrimination on the generated image. On the contrary, we perform fine-grained part discrimination on the image in the process of discriminating the generated image. Through these methods, we have significantly improved the visual quality and quantitative evaluation indicators for the generated images.

\section{Related Works}

Since generative adversarial network(GAN)\cite{goodfellow2014generative} was proposed in 2014, it has been the most widely used and best-performing base model in image generation. The GAN model consists of two competitors: a generator and a discriminator. Generator G takes the noise vector z sampled from the standard Gaussian distribution as input and then upsamples it into a generated image via a deep neural network. The discriminator D takes the real picture and the generated picture as input, and makes a decision on its authenticity, and outputs a real number from 0 to 1 to indicate the probability that the input image is a real image.

Although the generative adversarial network has made a huge breakthrough in image generation tasks, it only learns the mapping relationship from noise space to image. We can't input the semantics we specify to generate specific images. Conditional generative adversarial network\cite{mirza2014conditional} solves this problem. We can enter specific semantic information to constrain the visual representation of the resulting image. For example, when we generate handwritten digital images, we can input 0 to 9 digital semantic information before generation and then generate corresponding handwritten digital images that are realistic.

In order to further generate high-resolution and realistic images that can deceive human, we divide the generation of images into several stages. Except the initial stage of the total generating stages, the semantics and the output of the previous stage are added as inputs, and the resolution is gradually increased. At the same time, the output of each generation stage uses an independent discriminator to determine the realism and matching degree to the input semantics. This is StackGAN\cite{zhang2017stackgan}.

AttnGAN\cite{xu2018attngan} adds attention mechanism and matching mechanism across text and image on the stacked structure, and achieves certain breakthroughs in image generation tasks under fine-grained semantic constraints.

AttnGAN is one of the great achievements in the field of image generation in recent years, and its model framework has been widely used for further research. However, at present, the quality of image generation model under fine-grained semantic constraints after AttnGAN has not made great progress, and there is still the problem that the visual quality of generating pictures is far from satisfying human senses.

\section{Image Synthesis Model with Fine-grained Semantic Constraint}

The overall structure of the image generation network under the fine-grained semantic constraints proposed in this paper is shown in the figure \ref{fig:arch}. The improved part of our generation model is: \\
1. The semantic input module with richer and more accurate semantic constraint information. \\
2. A mask encoding and discriminating module that adds additional prior information.\\
3. A more accurate and fine-grained realism discriminating module. 

In the following sections of this section, we will elaborate the motivations and theories of our modules, and present our full objective function.

\subsection{Fine-grained Semantic Constraint}

\begin{figure}
	\begin{center}
		\includegraphics[width=0.9\linewidth]{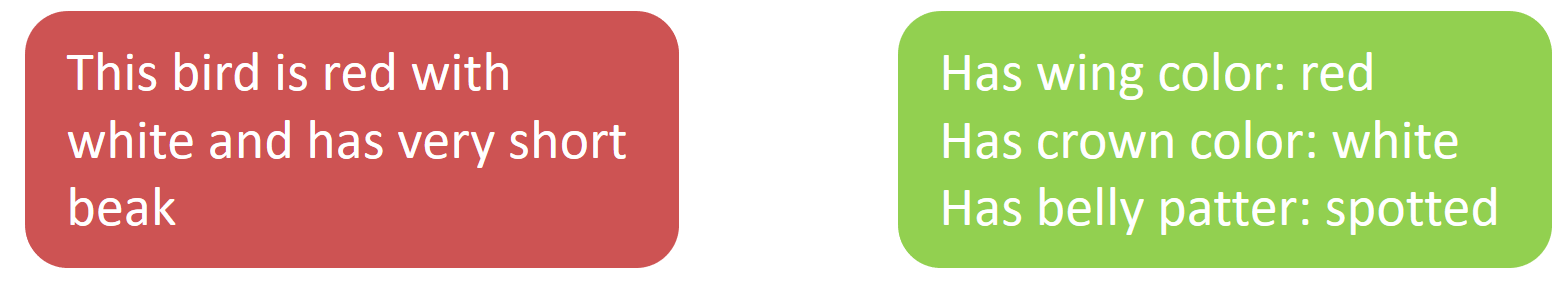}
	\end{center}
	\caption{Left: input in other models. Right: Our input.}
	\label{fig:attri}
\end{figure}

\textbf{Semantic Constraint with More Information.} At present, most of image generation model takes the embedded vector of a short sentence as a semantic input. However, the semantic information contained in a textual phrase is limited and will be disturbed by irrelevant words. In order to obtain richer and more fine-grained semantic input, we re-select the embedded vector of the feature attribute as input, as shown in the figure \ref{fig:attri}.

With this change, we can provide richer semantic constraints for generating images. Such input is the basis for the image-generating algorithm under our fine-grained semantic constraints to greatly improve the visual quality.

\begin{figure}
	\begin{center}
		\includegraphics[width=0.9\linewidth]{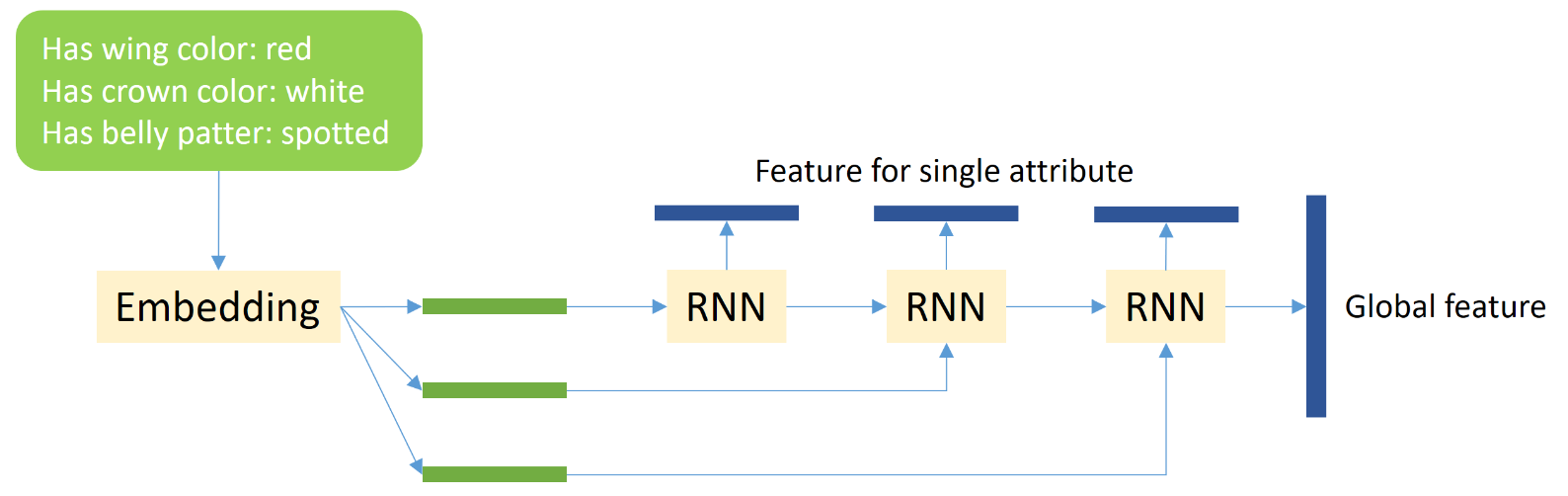}
	\end{center}
	\caption{Our encoding method to get global and local embedding of attribute.}
	\label{fig:rnn}
\end{figure}

Specifically, in our model, each attribute is encoded by the RNN like a word. We will use the final hidden layer state of the RNN as the global attribute feature vector, and the output vector by the RNN in each step will be a local attribute feature vector that is considered to correspond to an attribute, as shown in figure \ref{fig:rnn}.

\textbf{More Accurate Semantic Information.} Re-selection of attributes as input will undoubtedly help the image generation task under our fine-grained constraints, but it also introduces a new problem, that is, we need to accurately mark the training data set of the attribute. This kind of data set is very rare.

The CUB dataset is a dataset of bird images. Each instance in the dataset is labeled with 10 textual descriptions and 312-dimensional attribute vectors. These attribute labels are exactly the fine-grained semantic constraints we need. However, through careful investigation of the CUB dataset, we found that nearly 40\% of its attribute labeled did not match the image. 

\begin{figure}
	\begin{center}
		\includegraphics[width=\linewidth]{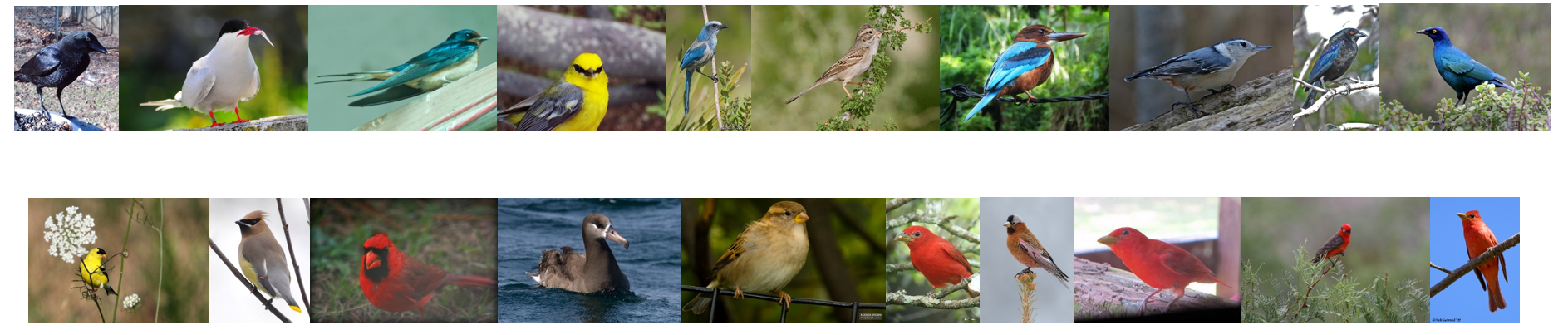}
	\end{center}
	\caption{Samples in CUB with blue(top) attribute or red(bottom) attribute.}
	\label{fig:sample}
\end{figure}

As shown in the figure \ref{fig:sample}, the first row of images is 10 randomly selected samples with blue wing attributes, and the second row of images is 10 randomly selected samples with red wing attributes.

Although the CUB dataset is labeled with attributes, it is certainly not a dataset that accurately labels the attributes. Nearly 40\% of the attributes of the dataset are not matched to the image. This is an inevitable problem with most data sets: there is always noise in manual labeling.

In order to enhance the correspondence between image features and attributes in datasets, we propose a manually labeled attribute denoising algorithm inspired by RANSAC\cite{fischler1981random} algorithm to clean the datasets marked by such noisy attributes.

\subsection{Extra Mask Prior}

\begin{figure}
	\begin{center}
		\includegraphics[width=\linewidth]{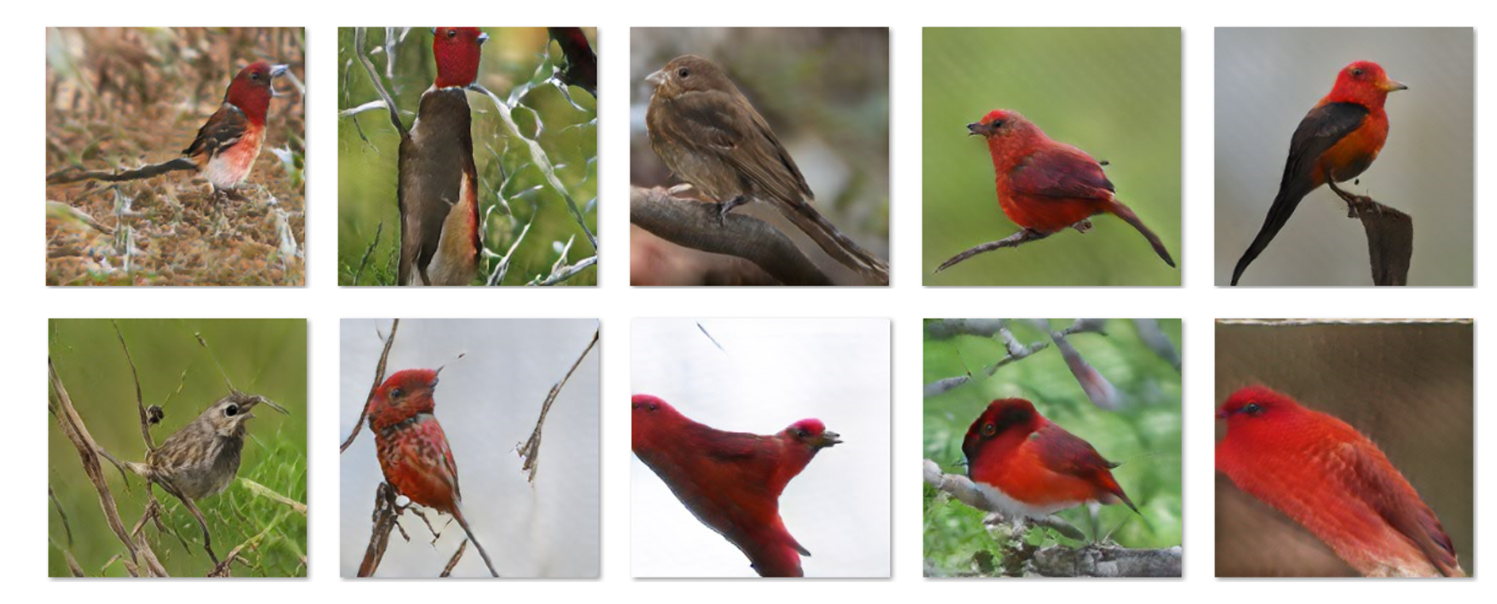}
	\end{center}
	\caption{Images randomly generated by AttnGAN.}
	\label{fig:attngansample}
\end{figure}

At present, many image generation models can generate high-quality images, but there are always strange images that are incomprehensible in the results. As shown in the figure \ref{fig:attngansample}, the image in the figure is generated by AttnGAN, and the images is generated by the same text. We can find that only a few images conform to the human visual sense of birds.

\begin{figure}
	\begin{center}
		\includegraphics[width=\linewidth]{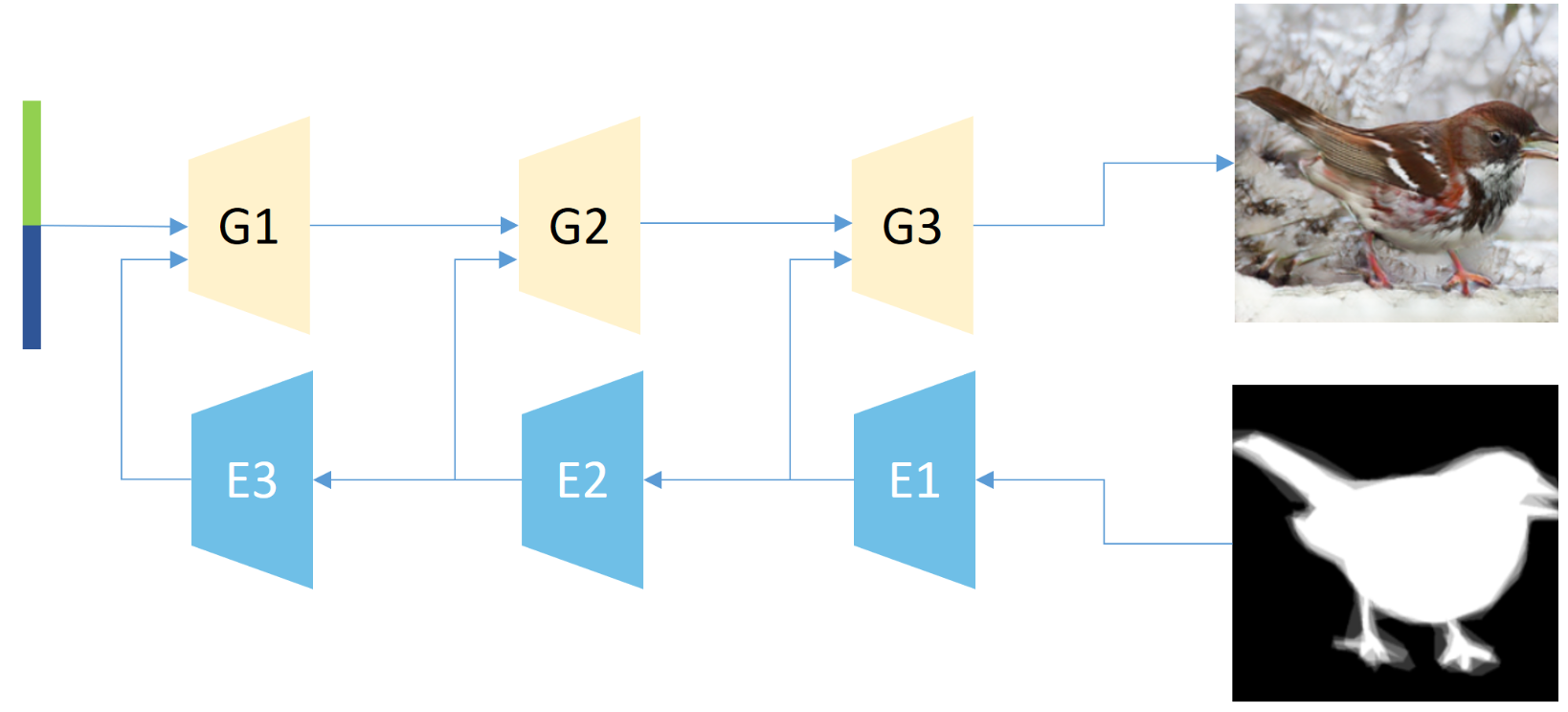}
	\end{center}
	\caption{Mask encoding module.}
	\label{fig:maskencoder}
\end{figure}

To solve this problem, we added mask constraints in the image generation process. Considering the stacked structure of the overall generated network, we use a U-NET structure as shown in figure \ref{fig:maskencoder} to encode the mask into different generation stages. We use less convolution and retain the number of high-resolution generation stages. More mask features of high-frequency information use mask features with more convolutions and more low-frequency information in the low-resolution generation phase.

Simply adding mask constraints is not enough. We also need to add corresponding objective functions so that the generation network learns how to make mask constraints play a role in the process of training. The mask loss proposed in this paper is as follows: 

\begin{align}
	\mathcal{L}_{mask}^{i} &= \mathbb{E}_{x \sim p_{data}(x)}[log D_{mask}^i(x \odot mask)] \\
	&+ \mathbb{E}_{z \sim \mathcal{N}(0,1)}[log(1-D_{mask}^i(G^i(z,c) \odot mask))]
\end{align}

\noindent where $i$ denotes the $i^{th}$ generation phase and $\odot$ denotes the AND operation.

\begin{figure}
	\begin{center}
		\includegraphics[width=\linewidth]{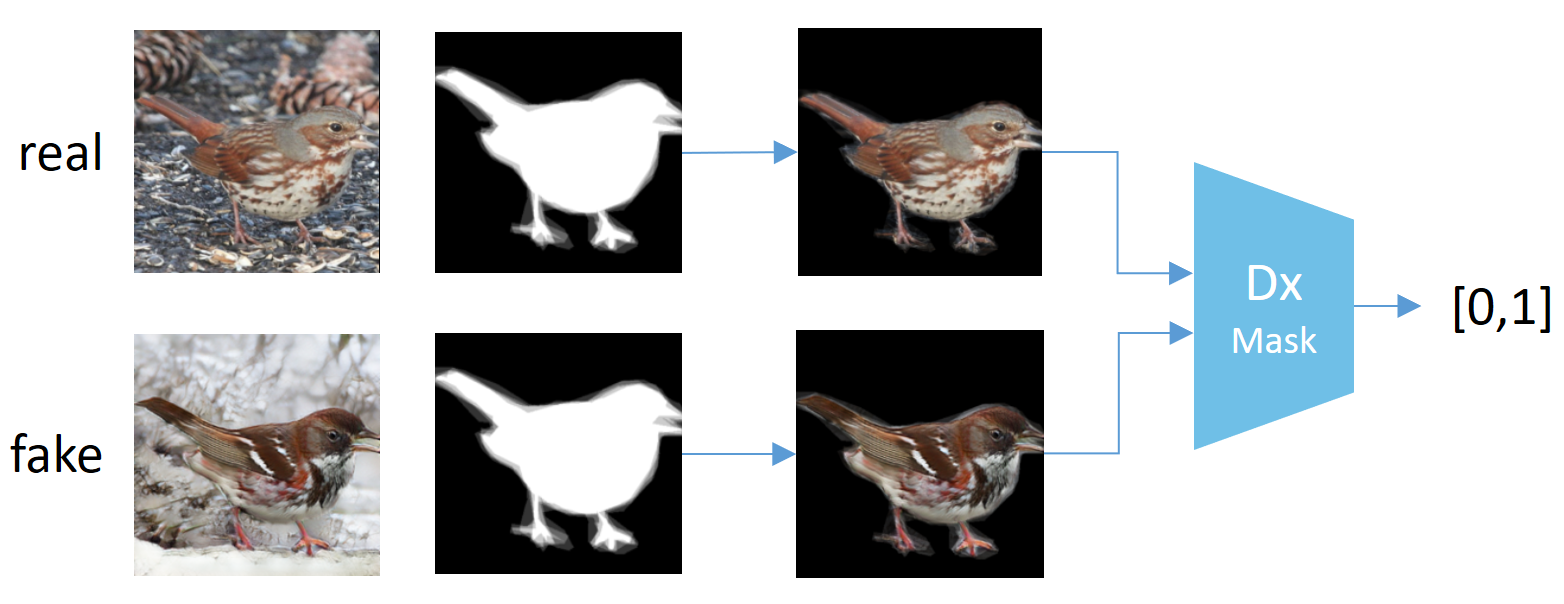}
	\end{center}
	\caption{Foreground discriminator.}
	\label{fig:maskD}
\end{figure}

As described in the equation, we will make an AND operation of the generated image, the real image and their corresponding masks, and crop the foreground of the image in the original image. This foreground image will be input into a discriminator to distinguish its realism. As shown in figure \ref{fig:maskD}, the discriminator's $x$ represents one such discriminator at each stage of generation.

By adding this objective function, the generation network will learn how to use the input mask features and further increase the authenticity of the foreground in the generated image.

\subsection{Fine-grained Part-Discriminator}

The popular image generation model often only makes overall discrimination on the generated image, and in order to enhance the visual quality of the details of the generated image, we make more fine-grained part-discrimination for images.

\begin{figure}
	\begin{center}
		\includegraphics[width=.8\linewidth]{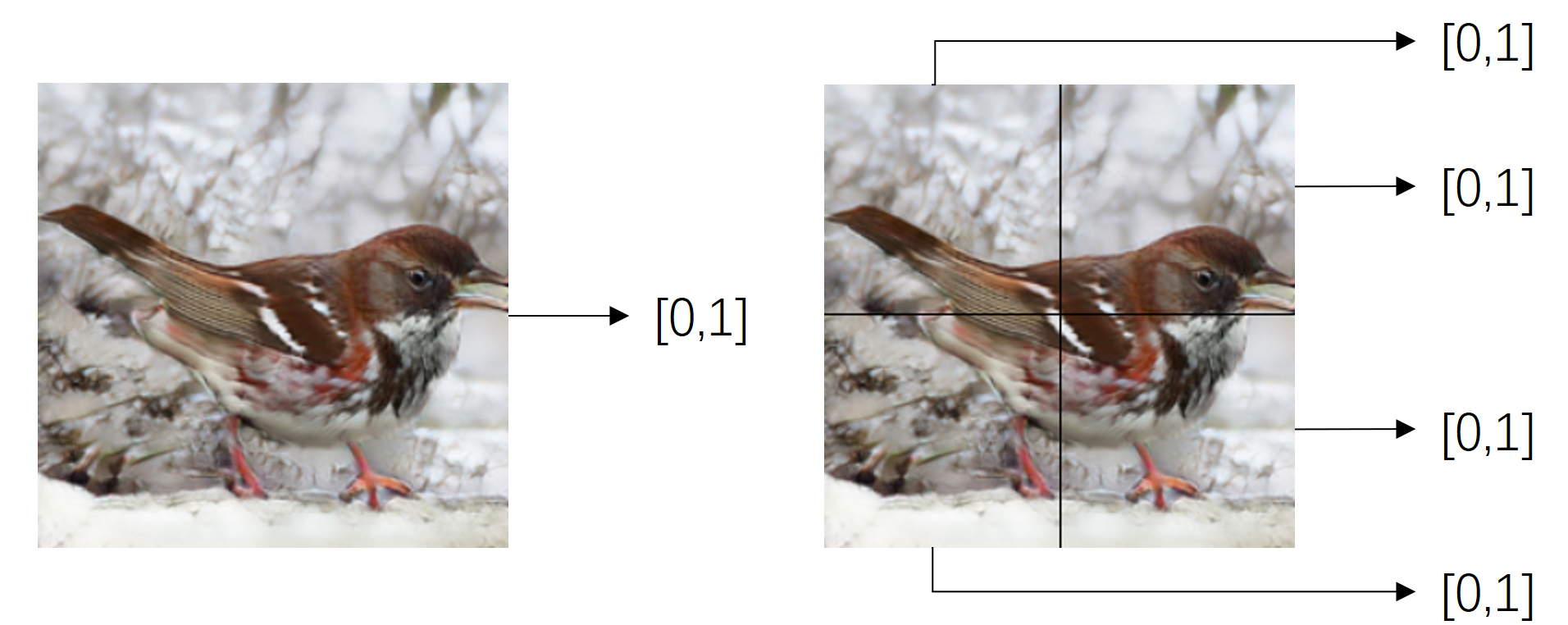}
	\end{center}
	\caption{Part discriminator.}
	\label{fig:partD}
\end{figure}

As shown in the figure \ref{fig:partD}, we divide the image into several regions inspired by HDGAN\cite{ledig2017photo}. In the discriminator, we not only calculate the overall realism of an image, but also calculate a local realism for each region. We will add this finer-grained discriminating structure when we judge the realism of the whole picture and the foreground picture. Therefore, our discriminator’s objective function for realism is divided into two parts, namely: 

\begin{equation}
	\begin{aligned}
		\mathcal{L}_{all}^{i} &= \mathbb{E}_{x \sim p_{data}(x)}[log D_{ori}^i(x)] \\
		&+ \mathbb{E}_{z \sim \mathcal{N}(0,1)}[log(1-D_{ori}^i(G^i(z,c)))]
	\end{aligned}
\end{equation}

\noindent and

\begin{equation}
	\begin{aligned}
		\mathcal{L}_{part}^{i} &= \frac{1}{4} \sum_{j=1}^4 (\mathbb{E}_{x \sim p_{data}(x)}[log D_{ori}^i(x_j)] \\
		&+ \mathbb{E}_{z \sim \mathcal{N}(0,1)}[log(1-D_{ori}^i(G^i(z,c)_j))])
	\end{aligned}
\end{equation}

\noindent $x_j$ and $G(*)_j$ respectively represent the $j^{th}$ region of the corresponding image.

This idea of part discrimination will greatly improve the realism of the generated image, thus improving the generation quality of the generated model.

\subsection{Full Objective Function}

In addition to the aforementioned mask decision function and the realistic decision function, we also need a conditional discriminant function to constrain the features of the generated image to match our semantic input.

\begin{equation}
	\begin{aligned}
		\mathcal{L}_{condition}^{i} &= \mathbb{E}_{x \sim p_{data}(x)}[log D_{cond}^i(x|c_{match})] \\
		&+\mathbb{E}_{x \sim p_{data}(x)}[log(1-D_{cond}^i(x|c_{not\_match}))]
	\end{aligned}
\end{equation}

Our full objective function is:

\begin{equation}
	\begin{aligned}
		\mathcal{L} = \sum_{i=1}^{3} (\mathcal{L}_{all}^i + \mathcal{L}_{part}^i + \mathcal{L}_{mask\_all}^i &+ \mathcal{L}_{mask\_part}^i \\
		&+ \mathcal{L}_{condition}^i)
	\end{aligned}
\end{equation}

In order to further fine-grain the degree of matching between the generated image and our semantic input, in the process of training the generator, we also add the cross-modality similarity model (DAMSM)\cite{xu2018attngan} loss proposed by AttnGAN. Therefore, the objective while training the generator is:

\begin{equation}
	\begin{aligned}
		\underset{G}{min} \  \mathcal{L} + \lambda\mathcal{L}_{DAMSM}
	\end{aligned}
\end{equation}

The objective while training the discriminator is:

\begin{equation}
	\begin{aligned}
		\underset{D}{max} \  \mathcal{L}
	\end{aligned}
\end{equation}

\section{Experiments}

\subsection{Data set}

Since many image generation methods have been tested and evaluated on the CUB dataset in recent years, our method will also be tested on the CUB dataset in order to facilitate comparison with previous work. Our experimental configuration on CUB is shown in the table \ref{table:cubset}.

There are 200 bird images in the CUB\cite{wah2011caltech} dataset, and there are about 60 bird images in the corresponding category in each category, for a total of 11,788. In our experiment, we will select a total of 8855 images from 150 classes and the mask corresponding to each image and the 312-dimensional attribute annotation as training samples. In the course of the test, we use the properties and masks of 2933 samples from the 50 classes that we have not learned during the training process as input to generate the corresponding bird image.

\begin{table}
	\begin{center}
		\begin{tabular}{|l|c|c|}
			\hline
			Split & Train & Test \\
			\hline\hline
			\# of class & 150 & 50 \\
			\hline
			\# of instance & 8855 & 2933 \\
			\hline
		\end{tabular}
	\end{center}
	\caption{Experiment Setting for CUB}
	\label{table:cubset}
\end{table}

\subsection{Quantitative Evaluation Method}

Similar to the evaluation methods of others image generation algorithms in recent years, we also use Inception Score\cite{salimans2016improved} as our evaluation method for generating models.

To calculate the generated image Inception Score of our generated model on the test set, we used a standard Inception-V3 classification network pre-trained on the original image of the test set to calculate the classification probability distribution and use the canonical Inception Score evaluation algorithm to calculate the score of the results of every model in our experiments.

\subsection{Details}

Since the AttnGAN model integrates useful models and modules for generating networks, stacked generation structures, and attention mechanisms in recent years, our network structure will use AttnGAN as the Baseline model, and add the modules proposed in this paper.

In the training of the model in this paper, we will pre-train a cross-modal matching model (DAMSM) and its required cyclic neural network (RNN) according to the strategy in AttnGAN. In the training of GAN, we choose Adam optimizer as our optimization method, and set the parameters of the optimizer to LR=0.0002, $\beta_1=0.5$, $\beta_2=0.999$. In our adversarial training process, we first optimize the discriminator once, then optimize the generator again, so that they reciprocate to make them tend to Nash equilibrium.

\subsection{Results and Analysis}

Before the training begins, we optimize the labeled attribute of the CUB dataset using the manual  attribute denoising algorithm described in Section 3.1. In the optimization process, we use the output before the last fully connect layer of the pre-trained ResNet-101 as the feature of the image, and use the SVM to perform the classification described in the manual attribute denoising algorithm.

\begin{figure}
	\begin{center}
		\includegraphics[width=\linewidth]{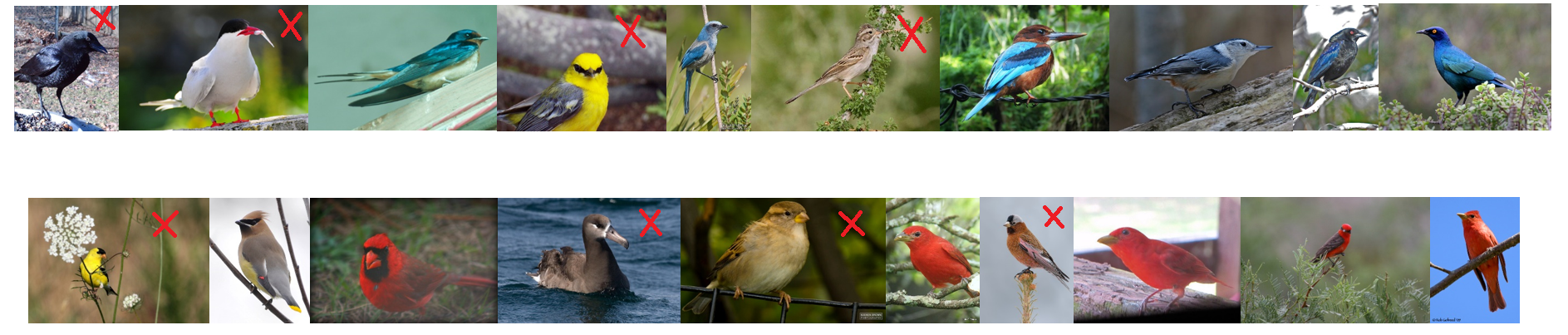}
	\end{center}
	\caption{Cleaned images according to optimized attributes.}
	\label{fig:optattri}
\end{figure}

With the optimized attribute, we searched the images selected in the figure one by one, and found that the label correspondence of the blue wing (top) and red wings (bottom) attributes is greatly improved, as shown in the figure \ref{fig:optattri}, where the images with red mark are no longer holding the wrong attribute.

We use the original AttnGAN as the Baseline, and then use the original CUB attributes and optimized attributes to train our networks. The comparison results in quantitative evaluation are shown in the table \ref{table:attri}.

\begin{table}
	\begin{center}
		\begin{tabular}{|l|c|}
			\hline
			Model & IS \\
			\hline\hline
			Baseline & 4.36$\pm$.03 \\
			\hline
			w. CUB attribute as input & 4.95$\pm$.07 \\
			\hline
			w. optimized attribute as input & \textbf{5.05$\pm$.06} \\
			\hline
		\end{tabular}
	\end{center}
	\caption{Model evaluation with attribute as input}
	\label{table:attri}
\end{table}

We can see that the attribute, which has more rich semantic constraints, can help the generative model to greatly improve the generating quality. Moreover, when the correspondence between the attribute semantics and the training picture is enhanced, the image generation quality is further improved.

After adding the mask encoding module and the foreground discriminating module as described in Section 3.3 and the fine-grained discriminating module as described in the section, the quantitative improvement results of the generated quality are as shown in the table \ref{table:extramodule}.

\begin{table}
	\begin{center}
		\begin{tabular}{|l|c|}
			\hline
			Model & IS \\
			\hline\hline
			w. optimized attribute & 5.05$\pm$.06 \\
			\hline
			w. mask module & 5.56$\pm$.08 \\
			\hline
			w. part discriminator & \textbf{5.91$\pm$.06} \\
			\hline
		\end{tabular}
	\end{center}
	\caption{Model evaluation with extra modules}
	\label{table:extramodule}
\end{table}

Through the incremental experimental strategy and corresponding quantitative evaluation of the generated quality, we prove that the mask constraint module and the fine-grained part discriminating module proposed in this paper play a key role in improving the quality of image generation.

\begin{figure}
	\begin{center}
		\includegraphics[width=\linewidth]{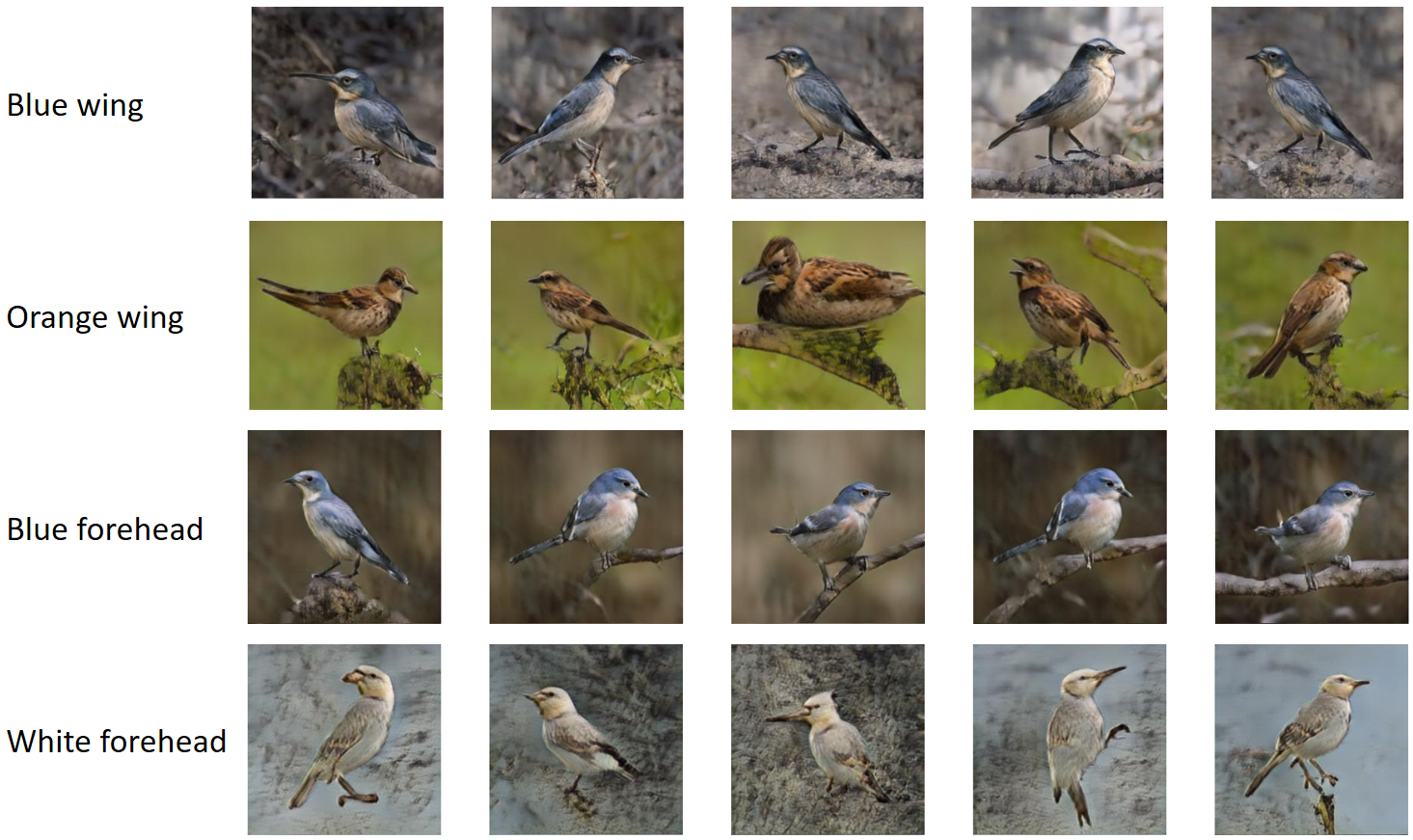}
	\end{center}
	\caption{Images generated by our model with only one attribute.}
	\label{fig:one2img}
\end{figure}

In addition, although the optimized attribute shown in the table \ref{table:attri} only help the generated model to increase by 0.1 on the quantitative evaluation, it actually improves a lot of consistency between the generated image features and the semantic constraints. As shown in the figure \ref{fig:one2img}, we generate several images using a single attribute, and the generated images were constrained by the corresponding attributes.

At the same time, we can compare the graph with the graph, we can find that the images generated by our model have good visual quality, and there is no incomprehensible bird image.

\begin{figure}
	\begin{center}
		\includegraphics[width=.9\linewidth]{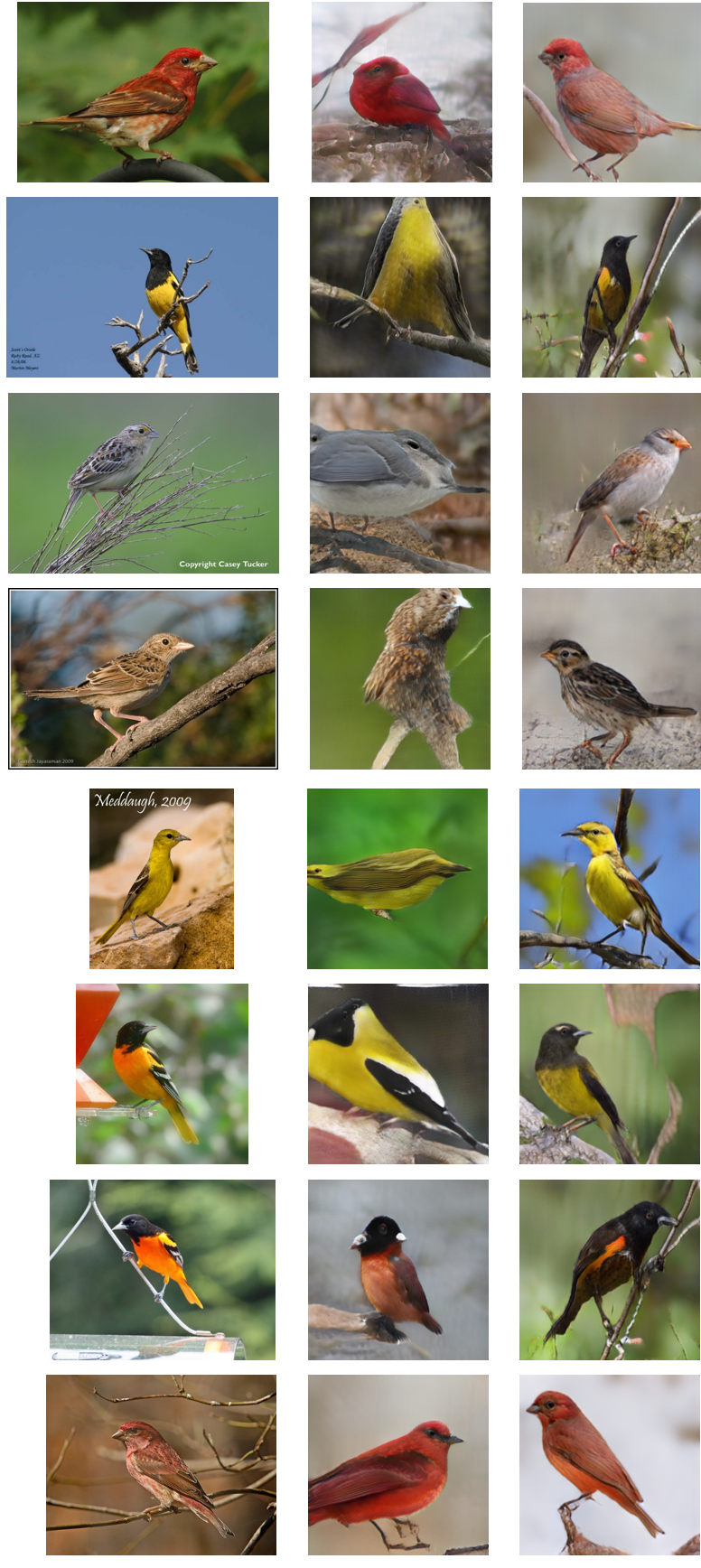}
	\end{center}
	\caption{Ground truth(left) and images generated by AttnGAN(mid) and our model(right) with label of ground truth.}
	\label{fig:attri2img}
\end{figure}

We also compare the image generated by our model with the image generated by the previous research, as shown in the figure \ref{fig:attri2img}, the first column is the real image, the second column is the image generated by AttnGAN, and the third column is generated by our model. We can clearly find that the images generated by our model have more fine-grained consistency and better visual quality than AttnGAN's results.

\section{Conclusion}

In this paper, through the research of various generation models based generative methods, we find a way to further solve the image generation problem under fine-grained semantic constraints. For the quantitative evaluation indicator Inception Score, the proposed model is 35\% higher than the latest work. At the same time, the image generation algorithm based on fine-grained semantic constraints proposed in this paper has reached the leading level in human visual perception.

\bibliographystyle{named}
\bibliography{bib}

\end{document}